\pdfoutput=1

\documentclass[11pt]{article}

\usepackage[final]{acl}

\usepackage{tabulary}
\usepackage{booktabs}
\usepackage{mdframed}

\usepackage{tabularx} 
\usepackage{lipsum} 
\usepackage{geometry} 

\usepackage{times}
\usepackage{latexsym}

\usepackage[T1]{fontenc}

\usepackage[utf8]{inputenc}

\usepackage{microtype}

\usepackage{inconsolata}

\usepackage{graphicx}
\usepackage{glossaries}
\usepackage{algorithm}
\usepackage{algpseudocode}

\usepackage{makecell}
\usepackage{multirow}

\newacronym{llm}{LLM}{Large Language Model}
\newacronym{ml}{ML}{Machine Learning}
\newacronym{kd}{KD}{Knowledge Distillation}
\newacronym{pgkd}{PGKD}{Performance-Guided Knowledge Distillation}
\newacronym{plm}{PLM}{Pre-trained Language Model}
\newacronym{ai}{AI}{Artificial Intelligence}
\newacronym{sla}{SLA}{Service-Level Agreement}
\newacronym{nlp}{NLP}{Natural Language Processing}
\newacronym{sota}{SOTA}{state-of-the-art}

\title{Performance-Guided LLM Knowledge Distillation for Efficient Text Classification at Scale}

\author{
   \textbf{Flavio Di Palo}\thanks{Equal contribution.} \hspace{1cm}
   \textbf{Prateek Singhi}\footnotemark[1] \hspace{1cm}  
   \textbf{Bilal Fadlallah} 
 \\
   Amazon
 \\
 \texttt{\{paloflav, snghips, bhf\}@amazon.com}
}

\begin{document}
\maketitle
\begin{abstract}
Large Language Models (LLMs) face significant challenges at inference time due to their high computational demands. To address this, we present Performance-Guided Knowledge Distillation (PGKD), a cost-effective and high-throughput solution for production text classification applications. PGKD utilizes teacher-student Knowledge Distillation to distill the knowledge of LLMs into smaller, task-specific models. PGKD establishes an active learning routine between the student model and the LLM; the LLM continuously generates new training data leveraging hard-negative mining, student model validation performance, and early-stopping protocols to inform the data generation. By employing a cyclical, performance-aware approach tailored for highly multi-class, sparsely annotated datasets prevalent in industrial text classification, PGKD effectively addresses training challenges and outperforms traditional BERT-base models and other knowledge distillation methods on several multi-class classification datasets. Additionally, cost and latency benchmarking reveals that models fine-tuned with PGKD are up to 130X faster and 25X less expensive than LLMs for inference on the same classification task. While PGKD is showcased for text classification tasks, its versatile framework can be extended to any LLM distillation task, including language generation, making it a powerful tool for optimizing performance across a wide range of AI applications.
\end{abstract}

\section{Introduction}
Successful research in \gls{ai} and \glspl{llm} brought significant interest to the industrial applications of \glspl{llm}. \glspl{llm} mainly refer to Transformer-based~\citep{Attention_is_all_you_need} neural language models that contain tens to hundreds of billions of parameters, which are pre-trained on massive text data, such as Mistral 7B~\citep{Jiang2023Mistral7}, LLaMA~\citep{touvron2023llama}, and GPT-4~\citep{openai2024gpt4}. Compared to smaller \glspl{plm} like BERT~\citep{devlin-etal-2019-bert} and GPT-2~\citep{Radford2019LanguageMA}, \glspl{llm} not only have significantly larger model sizes but also demonstrate stronger language understanding and generation capabilities.
It has been demonstrated that \glspl{llm} can achieve considerable performance with limited task-specific annotated data across several domains, including natural language understanding, question answering, and code generation~\citep{Minaee2024LargeLM}. 
In the text classification domain, \glspl{llm} like GPT-3 outperforms ad-hoc \gls{sota} models on several benchmarks by just using few-shot prompting techniques~\citep{sun-etal-2023-text}. 

While the utility of \glspl{llm} is evident, their application at scale is challenged by high inference latency and cost. \glspl{llm} are unsuitable for production environments where \glspl{sla} are strict and scalability is essential. Furthermore, many \gls{nlp} tasks in the industry only require a subset of the \glspl{llm} capabilities, e.g., Intent Detection, a multi-class classification problem that can be solved by fast and inexpensive \gls{plm} and does not require language generation capabilities. 
Furthermore, utilizing \glspl{llm} for text classification restricts the nuanced output information derived from fine-tuning a \gls{plm}. \glspl{plm}, when equipped with a classification head, produce output vectors with predicted probabilities that reflect the model's certainty about the classification results. However, this level of detail is not easily attainable from \glspl{llm}. \glspl{llm}, instead, generate plain text that requires parsing the model output and handling hallucinations, complicating production systems even further for simple tasks like text classification.

Considering the challenges presented by \glspl{llm} along with the effectiveness they bring, we propose a novel method called \gls{pgkd}. 
The proposed methodology aims to enhance the performance of smaller \glspl{plm}, which already meet inference \gls{sla} and cost constraints, by using \glspl{llm} as teachers during training.

\section{Related Work}
\gls{kd}, introduced by~\citep{hinton2015distilling}, is a promising technique to transfer the capabilities of complex and high-maintenance models to more compact and efficient student models. The core idea is to train a lean student model to mimic the soft probabilities generated by a more complex and costly teacher model. 

Using \glspl{llm} for ground truth generation and as \gls{kd} teacher has found success in multiple tasks~\citep{chiang2023large,Gilardi_2023,chan2023chateval}. 
Recent research~\citep{gu2024minillm} has explored novel \gls{kd} with open-source \glspl{llm} and loss functions explicitly tailored for teacher-student distillation. The same work also summarizes the two commonly applied categories of \gls{kd}: white-box \gls{kd}~\citep{Gou_2021}, where the teacher parameters are available to use for model distillation, and black-box \gls{kd}, where only the teacher predictions are accessible. black-box \gls{kd} is less restrictive in terms of structural requirements for teacher models and student models, and can use closed-source teacher models (such as ChatGPT API); additionally, it does not require the private deployment of teacher models and does not present challenges in custom loss function convergence.

Most recent black-box \gls{kd} methods mainly focus on \gls{llm} data generation or augmentation.~\citep{lou2023universal} proposed AugGPT, a text data augmentation approach based on ChatGPT that rephrases each sentence in the training samples into multiple conceptually similar but semantically different samples. ZeroGen~\citep{ye2022zerogen} is a flexible and efficient zero-shot learning method, providing insights on data-free, model-agnostic knowledge distillation. ZeroShotDataAug~\citep{ubani2023zeroshotdataaug} investigates ChatGPT-generated synthetic training data to augment low-resource datasets, outperforming existing data augmentation approaches, and explores methodologies for evaluating the quality of the generated augmented data. SunGen~\citep{Gao2022SelfGuidedND} optimizes synthetic data generation in black-box knowledge distillation by employing adversarial sampling and iterative refinement to create diverse, challenging data that aligns with the teacher model's knowledge boundaries.

Despite these advancements, the potential of \glspl{llm} in \gls{kd} extends beyond mere data generation. Recent studies have explored dynamic interactions between teacher and student models at training time for effective \gls{kd}. 
\citep{liu2024evolving} proved the effectiveness of active learning for teacher-student distillation for \glspl{llm} and introduces the EvoKD framework. This framework is specifically tailored to enhance the capabilities of a student model within a few-shot learning scenario, where only a limited quantity of training data is available. 

Research on text classification using \gls{kd} has been focused on few-shot learning scenarios, such as 1-shot or 5-shot, where 1 to 5 training samples are provided to the base model, typically narrowing the task to binary classification. While this focus offers valuable insights for theoretical exploration, it does not adequately represent the complexity of text classification tasks encountered in industrial applications. In a typical industrial setting, classification challenges, such as intent detection, topic classification, and customer feedback analysis, involve a much larger number of categories. Moreover, in these settings, annotating hundreds to a few thousand samples is often feasible and economically viable, making the 1-shot or 5-shot evaluation unrealistic given the availability of annotated data.

\section{Methods}

\gls{pgkd} is an iterative \gls{kd} algorithm that aims to enhance the application of \gls{kd} in highly multi-class, sparsely annotated datasets common in industrial environments. 
The primary motivation for this work is to leverage active learning to strengthen the connection between the student and teacher models in the distillation process. \gls{pgkd} gives the teacher model direct and continued visibility into the learning status of the student model also making it generically extensible to wide variety of learning tasks. \gls{pgkd} addresses limitations of recent works in black-box KD, in particular: (1) EvoKD only shows results on binary-classification benchmark datasets, a setting that is infrequent in practical machine learning applications;
(2) ZeroGen only relies on the latent knowledge of the teacher and is not aware of the student's status; the teacher model in EvoKD is also not provided with an overview of the latest student model performance on all the available classes and only shows the \gls{llm} a few misclassified examples; while this approach is useful in the context of binary text classification, it can be limiting in a highly multi-class context where many classes might not be represented in the sample provided to the teacher. 
(3) In SunGen, distillation focuses on the capability of the teacher model to generate high entropy data; EvoKD too focusses on correctly classified and misclassified samples but ignores emphasizing the information that is contained in the "Hard Negative Samples", i.e., incorrectly classified samples which the student model is confident about. 
(4) previous works do not provide a clear termination criterion for the iterative algorithm but treat the number of epochs for the knowledge distillation process as a hyperparameter that needs to be tuned for each dataset.
\newline
The proposed \gls{pgkd} overcomes the limitation above by leveraging the following techniques.\\
\textbf{(1)~Gradual Evaluation Checks:} At each \gls{kd} step, the student model is evaluated on the validation set, and a report of student validation metrics like Accuracy, Precision, Recall, and F1 on each class is inserted in the teacher model prompt (Appendix~\ref{sec:appendix_prompt}). This approach allows the teacher model to observe the student's overall performance and helps the teacher model guide the direction of the optimization. For instance, a low accuracy on a specific class will signal the teacher model to generate more relevant data samples for that class; this makes the \gls{kd} process performance-aware and allows the teacher model to observe and directly optimize the student model performance. It also induces automatic class imbalance handling that is directly based on student performance. It is important to remark that in this process, only high-level validation metrics are shared with the \gls{llm}, but no validation sample is leaked. This ensures the mitigation of the crucial risk of over-fitting the validation set during the learning process.\\
\textbf{(2)~Hard Negative Mining:} Hard Negative samples are defined as the misclassified samples in the training set for which the student model is more confident in its decision. These samples are generally the most informative ones, being the closest to the student model decision boundary. \gls{pgkd} includes the Hard Negative samples from the previous training run in the teacher's prompt. The \gls{llm} evaluates the student model's performance deficiencies by accessing sentence patterns likely to result in errors.  The confidence of the student model is computed by considering the values from the classification head of the student model. This approach allows the teacher model to gain rich insights into which examples the student is struggling to learn and encourages the teacher to produce new samples that help overcome these crucial blind spots.\\
\textbf{(3)~Early Stopping:} To maximize the student model's distillation while preventing performance drift and overfitting \gls{pgkd} uses early stopping to control the overall validation loss. \gls{pgkd} returns the model having the lowest validation loss as the best model to be used for testing.

\begin{figure*}[!ht]
  \centering
    \includegraphics[width=\linewidth]{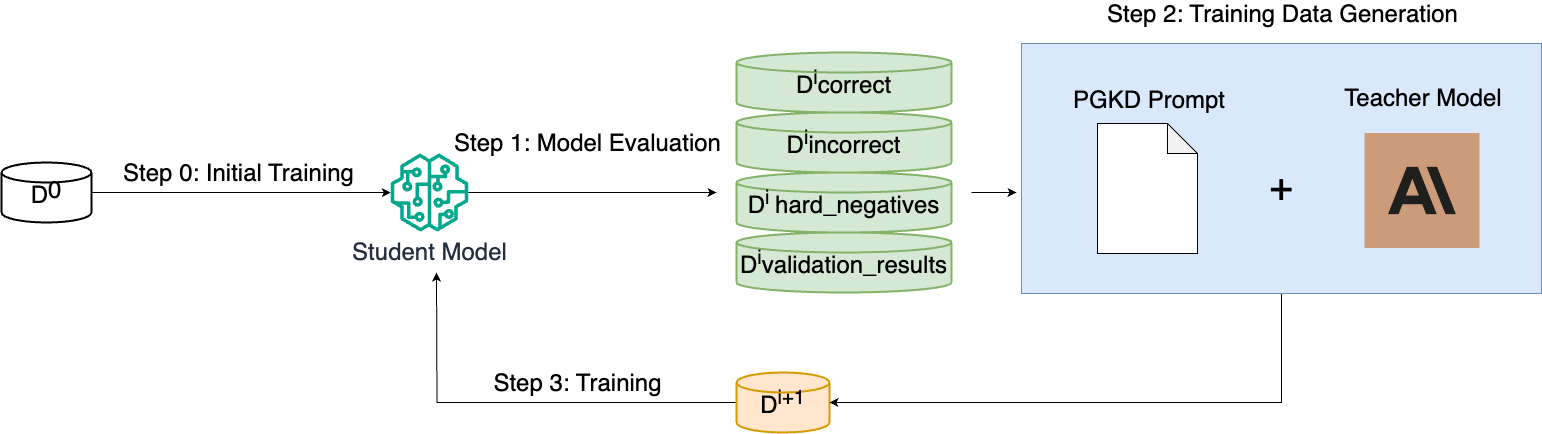}

  \caption{PGKD process, student model is initially trained on a set of labeled samples, then the \gls{kd} process starts. The \gls{llm} generates new training samples for the student model based on the student's correct/misclassified samples, hard negative samples, and a report of the student's validation metrics.}
  \label{fig:algorithm_scheme}
\end{figure*}

\begin{figure*}[!ht]
  \centering
    \includegraphics[width=\linewidth]{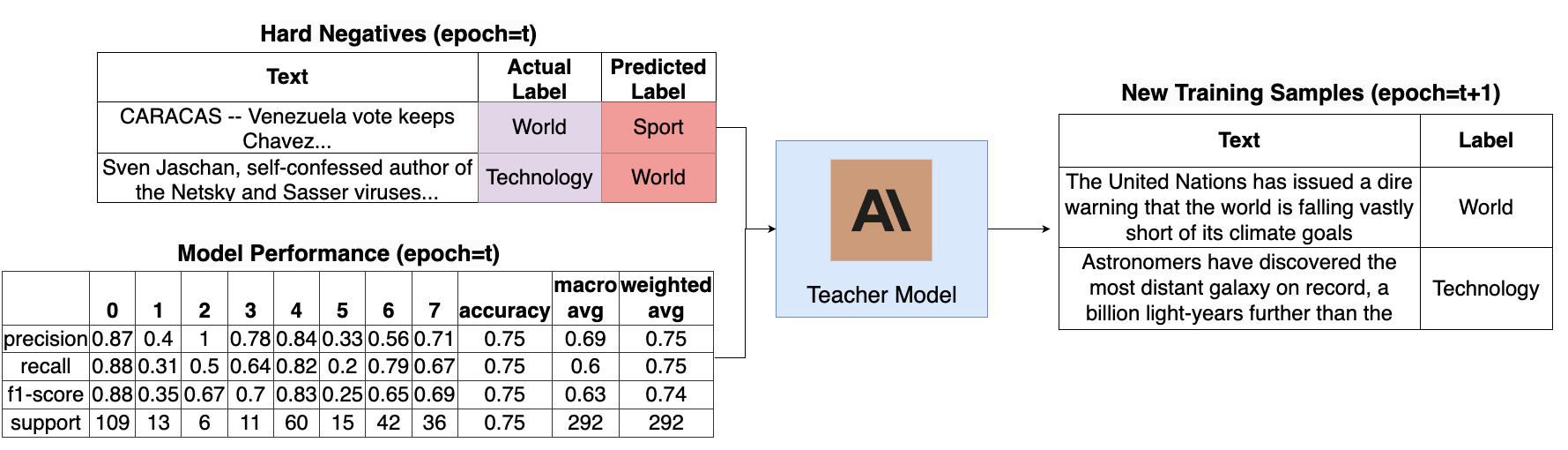}
  \caption{PGKD methods leverages Validation Metrics and Hard Negative samples from $epoch_{t}$ in the \gls{llm} prompt. The teacher model generates new samples to augment the training set for $epoch_{t+1}$.}
  \label{fig:validation_example}
\end{figure*}

The \gls{pgkd} methodology is detailed in Algorithm~\ref{alg:evokd} and a schema of the same is reported in Figure~\ref{fig:algorithm_scheme} and Figure~\ref{fig:validation_example}. 

\begin{algorithm*}[!ht]
 \caption{Performance-Guided Knowledge Distillation.}
\label{alg:evokd}
\begin{algorithmic}[1]
\Require $D^{0}$, $D^{val}$ \textit{num\_kd\_steps}, \textit{patience\_limit}, PGKD\_prompt
\State Initialize $model$; $model^{0} \leftarrow \text{train } model \text{ on } D^{0}$; $i \leftarrow 0$; $step \leftarrow 0$
\State $best\_validation\_loss \leftarrow \infty $; $val\_results \leftarrow ''$; $patience\_counter \leftarrow 0$; $history \leftarrow \{D^{0}\}$
\While{$i <= num\_kd\_steps$}
    \State $D_{incorrect}^{i}, D_{correct}^{i} \leftarrow \text{Find\_Correctly\_Classified\_And\_Misclassfied\_Examples}(model^{i}, D^{i})$
    \State $D^{i}_{hard\_negatives} \leftarrow \text{Find\_Hard\_Negative\_Examples}(model^{i}, D^{i}) $
    \State $D^{i+1}  \leftarrow \text{LLM}(D_{incorrect}^{i}, D_{correct}^{i}, D_{hard\_negatives}^{i}, val\_results, PGKD\_prompt)$
    \State Add $D^{i+1}$ to $history$

    \State $model^{i+1} \leftarrow train\ model\ on\ history$
    \State $new\_loss, val\_results \leftarrow evaluate(model^i, D^{val})$
    \If{$new\_loss > best\_validation\_loss $}: 
        \State $patience\_counter += 1$
        \If{$patience\_counter > patience\_limit $}: 
        \State $break$
        \EndIf  
    \Else
    \State $best\_validation\_loss \leftarrow  new\_loss$
    \State $model^{best} \leftarrow model^{i}$
    \EndIf
    \State $i \leftarrow i + 1$

\EndWhile
\State \Return $model^{best}$
\end{algorithmic}
\end{algorithm*}

The algorithm starts by initializing and training a baseline model on an initial dataset $D^0$ composed of a thousand annotated samples, divided into 80\% training and 20\% validation. The base model obtained after this training is denoted as $model_0$. \gls{pgkd} iteratively refines $model_0$ performance; during each iteration, the \gls{pgkd} algorithm identifies correctly classified $D^{i}_{correct}$ and misclassified examples $D^{i}_{incorrect}$, as well as hard negatives $D^{i}_{hard\_negatives}$. \gls{pgkd} then computes the model $val\_results$ on the validation set $D^{val}$ and leverages an \gls{llm} to generate new training data $D^{i+1}$ tailored to these findings according to $PGKD\_prompt$. $PGKD\_prompt$ includes dataset classification taxonomy for the specific task and a limited number of few-shot samples from the train set, prompt is reported in Appendix~\ref{sec:appendix_prompt}.
The \gls{pgkd} process runs for $num\_kd\_steps$ and if the validation loss does not improve consecutively beyond the $patience\_limit$, the algorithm terminates early to prevent overfitting. At the end of the \gls{pgkd} process, the best model on the validation set, $model^{best}$ is returned. 

 
\section{Datasets and Experiments}
Our experiments focus on four multi-class classification datasets: AG-news~\citep{agulli}, Yahoo Answers~\citep{Ardeshna2020}, Huffington Post~\citep{Misra2018}, and AMZN Reviews~\citep{Kashnitsky2018}, described in Table~\ref{tab:datasets_used}. 
\begin{table*}[ht!]
\resizebox{\linewidth}{!}{%
\noindent\begin{tabularx}{\textwidth}{>{\hsize=0.7\hsize}X>{\hsize=1.4\hsize}X>{\hsize=0.3\hsize}X>{\hsize=0.3\hsize}X>{\hsize=0.3\hsize}X}
\toprule
\textbf{Dataset} & \textbf{Description} & \textbf{\#Training Samples} & \textbf{\#Testing Samples} & \textbf{\#Classes} \\
\midrule
AG-News & Categorize news articles gathered from more than 2000 news sources in four categories (World, Sports, Business, Sci/Tech). & 120,000 & 7,600 & 4 \\ 
\midrule
Yahoo Answers & Categorizes answers from the popular website `Yahoo Answers' into ten different categories. & 1,400,000 & 60,000 & 10 \\
\midrule
Huffington Post & Categorizes Huffington Post news headlines from 2012–2022 into forty-one different categories such as politics, wellness, entertainment, and travel. & 160,000 & 40,000 & 41 \\
\midrule
AMZN Reviews & Identifies Amazon Product categories based on the review title and text for that product.    ``level 3'' classes present in the training sample used are considered targets for the classification task. & 40,000 & 10,000 & 335 \\
\bottomrule
\end{tabularx}}
\caption{Detailed description of the datasets used for the PGKD experiments. Each dataset varies significantly in the number of classes, training samples, and testing samples, reflecting a broad range of classification challenges.}
\label{tab:datasets_used}
\end{table*}
The datasets cover a wide range of number of classes, from 4 to 335. These experiments aim to study the \gls{pgkd} performance at the variation of the number of classes in the dataset. The base \gls{plm} model used for \gls{pgkd} is BERT-base model, as done in~\citep{liu2024evolving}; the model has been fine-tuned with a classification head using categorical cross-entropy leveraging the Hugging Face library\footnote{https://huggingface.co/docs/transformers/model\_doc/bert}. A sample of 1000 annotated data points is used for $model\_0$ training and validation (80\% training, 20\% validation). The teacher model for the experiments is Claude-3 Sonnet accessed via AWS Bedrock\footnote{https://aws.amazon.com/bedrock/}. Early stopping is applied on the validation loss for the initial BERT-base model and the \gls{pgkd} methodology is then applied to the resulting model. For robustness of the provided results, 5 different training and validation sets of 1000 samples are chosen; results are averaged across these 5 samples. 
For BERT-base $model\_0$ fine-tuning, the following parameters have been set: maximum sequence length of 512, batch size of 64, fine-tuned the model for 30 epochs with a learning rate of \(2 \times 10^{-5}\), and patience parameter for early stopping at 5.
\gls{pgkd} fine-tuning is tasked to produce 32 new samples at each iteration and the prompt takes as input 16 samples from the training set as few-shot samples used to guide the \gls{llm} data generation; the number of correct samples, incorrect samples, and hard negative samples is set to 16. The number of \gls{pgkd} epochs is set to 10 and \gls{pgkd} patience is set to 5.
The purpose of the experiments is to measure the performance lift registered by applying the \gls{pgkd} routine on top of the $model\_0$ BERT-base model for different datasets with a varying number of classes. 

\section{Results and Discussion}

\begin{figure*}[!ht]
  \centering
  \includegraphics[width=0.85\linewidth]{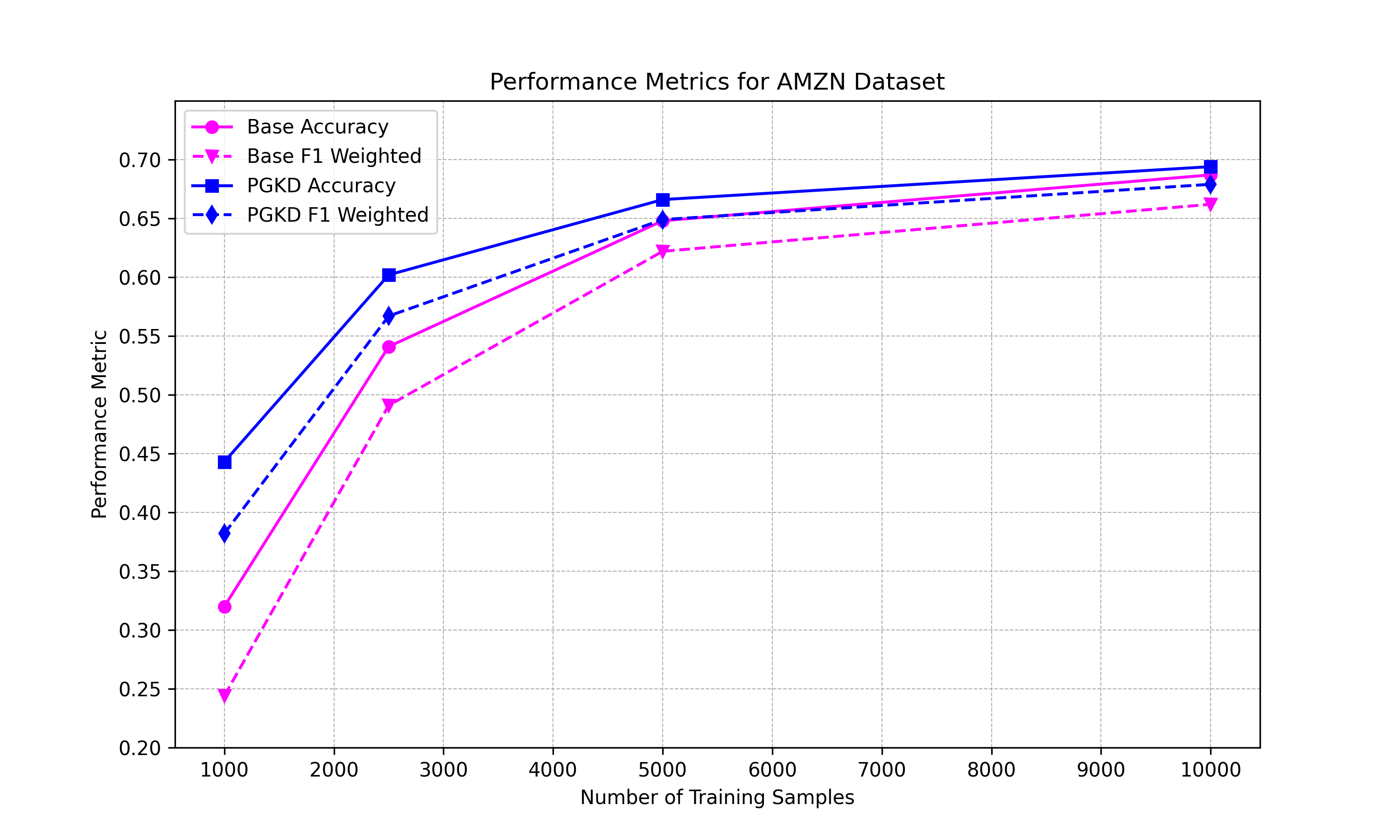}

  \caption{Performance of \gls{pgkd} across varying training set sizes on the AMZN Reviews dataset.}
  \label{fig:amzn_scaling}
\end{figure*}

\gls{pgkd} results are presented in Table~\ref{tab:model_results}.
\begin{table*}[!ht]
\centering
\resizebox{\linewidth}{!}{%
\begin{tabular}{ccccc} 
\toprule
 & \textbf{Method} & \textbf{Accuracy} & \textbf{Macro Avg. F1} & \textbf{Weighted Avg. F1} \\
 \midrule
\multirow{4}{*}{\begin{tabular}[c]{@{}c@{}}\textbf{\textbf{\textbf{\textbf{AG-news }}}}\\\textbf{\textbf{\textbf{\textbf{(\#classes = 4)}}}}\end{tabular}} & BERT-base & 0.884 $\pm$ 0.012 & 0.884 $\pm$ 0.013 & 0.884 $\pm$ 0.011 \\
 & BERT-base + PGKD & \textbf{0.895 $\pm$ 0.014} & \textbf{0.894 $\pm$ 0.018} & \textbf{0.894 $\pm$ 0.015} \\
 & Claude-3 (Zero-Shot) & 0.826 $\pm$ 0.020 & 0.821 $\pm$ 0.016 & 0.822 $\pm$ 0.012 \\
 & SOTA - Full shot~\citep{Yang2019XLNetGA} & 0.955 & - & - \\ 
\midrule
\multirow{4}{*}{\begin{tabular}[c]{@{}c@{}}\textbf{\textbf{Yahoo Answers }}\\\textbf{\textbf{(\#classes = 10)}}\end{tabular}} & BERT-base & 0.649 $\pm$ 0.013 & 0.657 $\pm$ 0.017 & 0.657 $\pm$ 0.015 \\
 & BERT-base + PGKD & \textbf{0.685 $\pm$ 0.015} & \textbf{0.688 $\pm$ 0.019} & \textbf{0.688 $\pm$ 0.011} \\
 & Claude-3 (Zero-Shot) & 0.680 $\pm$ 0.016 & 0.676 $\pm$ 0.018 & 0.676 $\pm$ 0.017 \\
 & SOTA - Full Shot~\citep{bert-xx} & 0.776 & - & - \\ 
\midrule
\multirow{4}{*}{\begin{tabular}[c]{@{}c@{}}\textbf{\textbf{Huffington Post }}\\\textbf{\textbf{(\#classes = 41)}}\end{tabular}} & BERT-base & 0.474 $\pm$ 0.011 & 0.214 $\pm$ 0.014 & 0.411 $\pm$ 0.013 \\
 & BERT-base + PGKD & \textbf{0.519 $\pm$ 0.012} & 0.330 $\pm$ 0.018 & \textbf{0.495 $\pm$ 0.020} \\
 & Claude-3 (Zero-Shot) & 0.442 $\pm$ 0.015 & \textbf{0.332 $\pm$ 0.019} & 0.435 $\pm$ 0.016 \\
 & SOTA - 5-way, 5-shot~\citep{contrast_net} & 0.653 & - & - \\ 
\midrule
\multirow{4}{*}{\begin{tabular}[c]{@{}c@{}}\textbf{\textbf{AMZN Reviews }}\\\textbf{\textbf{(\#classes = 335)}}\end{tabular}} & BERT-base & 0.320 $\pm$ 0.014 & 0.074 $\pm$ 0.019 & 0.244 $\pm$ 0.011 \\
 & BERT-base + PGKD & \textbf{0.443 $\pm$ 0.017} & 0.159 $\pm$ 0.012 & 0.382 $\pm$ 0.014 \\
 & Claude-3 (Zero-Shot) & 0.416 $\pm$ 0.012 & \textbf{0.364 $\pm$ 0.016} & \textbf{0.414 $\pm$ 0.013} \\
 & SOTA - 5-shot~\citep{zhang-etal-2022-mgimn} & 0.495 & - & - \\
\bottomrule
\end{tabular}
}
\caption{Comparison of average and standard deviation for Accuracy, Macro Average F1, and Weighted Average F1 scores for BERT-base and BERT-base enhanced with PGKD across various datasets with differing numbers of classes. Results are shown alongside Claude-3 zero-shot and \gls{sota} full-shot or few-shot methods as referenced in the literature.}
\label{tab:model_results}
\end{table*}
\gls{pgkd} exhibits a varying degree of effectiveness across the datasets.
\textbf{AG-news (4 classes)}: The \gls{pgkd} implementation yields only slight improvements in AG-news, a dataset with a reduced number of classes. The metrics show an increase in Accuracy from 0.884 to 0.895, Macro Average F1 from 0.884 to 0.894, and Weighted Average F1 from 0.884 to 0.894 compared to BERT-base. These modest gains likely reflect the already high baseline performance, which limits the scope for further significant enhancements through the \gls{pgkd} method.
\textbf{Yahoo Answers (10 classes)}: \gls{pgkd} shows a noticeable improvement in all measured metrics. Accuracy is enhanced from 0.649 to 0.685, Macro Average F1 from 0.657 to 0.688, and Weighted Average F1 from 0.657 to 0.688. This dataset’s moderate number of classes indicates that \gls{pgkd} is particularly effective in scenarios with intermediate complexity, utilizing the teacher model’s strengths more effectively.
\textbf{Huffington Post (41 classes)}: The application of \gls{pgkd} in the Huffington Post dataset, with a substantially higher number of classes, also leads to significant improvements. Accuracy improves from 0.474 to 0.519, Macro Average F1 from 0.214 to 0.330, and Weighted Average F1 from 0.411 to 0.495. These improvements underscore the benefits of \gls{pgkd} in managing more complex class structures and enhancing model performance.
\textbf{AMZN Reviews (335 classes)}: This dataset, featuring the largest number of classes, displays the most dramatic improvements with \gls{pgkd}. Accuracy significantly increases from 0.320 to 0.443, Macro Average F1 from 0.074 to 0.159, and Weighted Average F1 from 0.244 to 0.382. These results highlight \gls{pgkd}’s strong capability in handling extensive, complex class structures, particularly beneficial in contexts with scarce labeled data.
We observe a correlation between the number of classes and the improvement margin achieved through \gls{pgkd}. Datasets with a lower number of classes show negligible improvements, while datasets with a larger number of classes show significant performance gains. This suggests that the methodology is particularly suited for production applications where distinguishing between a large number of classes is challenging due to the complexity and sparsity of the annotated data; this class of problems is prevalent in many industrial \gls{ml} applications.
Performance of the current \gls{sota} methodology for each dataset and performance of Claude-3 Sonnet zero-shot classification are reported. The corresponding prompt for zero-shot classification is reported in Appendix~\ref{sec:appendix_prompt}. In all the reported use-cases, \gls{pgkd} outperformed zero-shot Claude-3 Accuracy improving the performance gap between BERT-base and current \gls{sota}; BERT-base + \gls{pgkd} outperforms Claude-3 zero-shot on Weighted Average F1 on all datasets except AMZN Reviews; BERT-base + \gls{pgkd} has superior Macro Average F1 results when compared with Claude-3 on two out of four datasets, with slightly lower results on the Huffington Post dataset. It is plausible that enhancing the prompt with complex few-shot techniques could elevate the \gls{llm} to match the current \gls{sota} performance, as reported in~\citep{sun-etal-2023-text}; however, further \gls{llm} prompt exploration is beyond the scope of this work.
While \gls{pgkd} is expected to be maximally useful when the number of annotated data is limited, it is interesting to evaluate performance gains obtained by \gls{pgkd} as the number of training samples increases. Figure~\ref{fig:amzn_scaling} presents the results for the AMZN Reviews dataset obtained at a varying number of training dataset sizes, with similar trends observed across all datasets; full results are documented in Appendix~\ref{appendix:scaling}. 
The \gls{pgkd} approach consistently surpassed the performance of the BERT-base model. Notably, as the number of training samples from the original dataset increases, the performance differential between \gls{pgkd} and the original model decreases. This phenomenon can be attributed to the diminishing returns of new data generation by the \gls{llm} as the volume of original dataset samples grows. Interestingly, while \gls{pgkd} performance improvement reduces with the increasing number of training samples, we can observe that the \gls{pgkd} provides the best results across all dataset considered for all training sample sizes. \gls{pgkd} demonstrates to not degrade model in any of the experiments conducted. 

\subsection{Comparative Analysis of Related Literature}
We compared \gls{pgkd} with other \gls{kd} strategies on two datasets: the IMDB Dataset~\citep{Lakshmi2018}, containing 2 classes, and the Inshorts News V7~\citep{Shashi2021}, containing 7 classes, as presented in~\citep{liu2024evolving}. 
Table~\ref{tab:evokd_comparison} reports the comparison between \gls{pgkd},
EvoKD and other baseline 1-shot classification methods from~\citep{liu2024evolving}. 
\begin{table}[!ht]
\centering
\resizebox{\linewidth}{!}{%
\begin{tabular}{lcc}
\toprule
\textbf{Method} & \textbf{IMDB} & \textbf{Inshorts} \\
\midrule
Full Shot & 0.949 & 0.970 \\
\midrule
BERT-base (n=1000) & 0.872 & 0.933 \\
BERT-base + PGKD (n=1000) & \textbf{0.908} & \textbf{0.943} \\
\midrule
No Augment (1-shot) & 0.583 & 0.640 \\
EDA (1-shot) & 0.618 & 0.677 \\
ZeroGen (1-shot) & 0.508 & 0.833 \\
SunGen (1-shot) & 0.576 & 0.810 \\
Gradual (1-shot) & 0.685 & 0.760 \\
AugGPT (1-shot) & 0.690 & 0.790 \\
\midrule
EvoKD (1-shot) & 0.798 & 0.851 \\
EvoKD + Init (1-shot) & 0.835 & 0.868 \\
\bottomrule
\end{tabular}}
\caption{Weighted Average F1 score comparison of Knowledge Distillation methods on English IMDB and Inshorts datasets. 1-shot methods reported from~\citep{liu2024evolving}.}
\label{tab:evokd_comparison}
\end{table}
The 1-shot approach discussed in~\citep{liu2024evolving} may be impractical for production environments, where access to larger datasets is the norm. The BERT-base model trained with 1000 samples achieves a weighted average F1 score of 0.872 on the IMDB dataset and 0.933 on the Inshorts dataset. Interestingly BERT-base already outperforms all the 1-shot methodologies reported in~\citep{liu2024evolving} even without \gls{pgkd}. \gls{pgkd} further improves upon this performance, achieving a weighted average F1 score of 0.908 on the IMDB dataset and 0.943 on the Inshorts dataset. The results demonstrate that \gls{pgkd} is able to improve BERT-base model performance in this context as well.

\subsection{Ablation Study}
Table~\ref{tab:ablation} reports the effectiveness of specific \gls{pgkd} components, proving the contribution of the proposed methodology over a general \gls{llm} active learning framework. 
\begin{table}[!ht]
    \centering
    \resizebox{\linewidth}{!}{%
    \begin{tabular}{lcccc}
        \toprule
        \textbf{Method} & \rotatebox{45}{\textbf{AG-news}} & \rotatebox{45}{\textbf{Yahoo A.}} & \rotatebox{45}{\textbf{Huffington}} & \rotatebox{45}{\textbf{AMZN}} \\
        \midrule
        PGKD & 0.895 & 0.685 & 0.519 & 0.443\\
        w/o Validation & 0.893 & 0.669 & 0.501  & 0.419 \\
        w/o Hard Negatives & 0.887 & 0.675 & 0.510  & 0.433 \\
        \bottomrule
    \end{tabular}}
    \caption{Average Accuracy on five training samples; \gls{pgkd} applied to BERT-base model trained with 1000 samples.}
    \label{tab:ablation}
\end{table}
Experiments have been carried out with the \gls{pgkd} methodology by selectively disabling the Validation Report and Hard Negatives features across different datasets. The removal of the Validation component (w/o Validation), which uses performance metrics to guide the distillation process, resulted in a decrease in accuracy across all datasets. Specifically, the accuracy decreased from 0.895 to 0.893 on AG-news, from 0.685 to 0.669 on Yahoo Answers, from 0.519 to 0.501 on Huffington Post, and more significantly from 0.443 to 0.419 on Amazon Reviews. This indicates that validation metrics information is useful in the distillation process by making it performance-aware; as expected, this feature is particularly relevant to datasets with a high number of classes. Removing Hard Negative Mining (w/o Hard Negatives), which incorporates challenging misclassified samples to refine the student model's decision boundaries, led to a more subtle drop in model performance. For instance, accuracy fell from 0.895 to 0.887 on AG-news, from 0.685 to 0.675 on Yahoo Answers, from 0.519 to 0.510 on Huffington Post, and from 0.443 to 0.433 on Amazon Reviews. This decline was more pronounced in datasets with complex classification tasks, highlighting the importance of hard negative samples in enhancing the robustness and accuracy of the student model in multi-class scenarios.

\subsection{Cost and Latency Benchmarking}
Inference latency and cost for \gls{pgkd} and \glspl{llm}\footnote{All \glspl{llm} are accessed via Amazon Bedrock: \url{https://aws.amazon.com/bedrock/}} are reported in Table~\ref{tab:inference_cost_latency}.
\begin{table}[!ht]
\centering
\resizebox{\linewidth}{!}{%
\begin{tabular}{lcc}
\toprule
\textbf{Method} & \textbf{Latency (s)} & \textbf{Cost (\$)} \\
\midrule
BERT-base + PGKD (CPU) & 21.45 & 0.0046 \\
BERT-base + PGKD (GPU) & 0.46 & 0.0107 \\
\midrule
Claude-3 zero-shot & 60.64 & 0.3867 \\
LLaMA 3 8B zero-shot & 58.05 & 0.0609 \\
\bottomrule
\end{tabular}}
\caption{Comparison of latency in seconds and cost in dollars for BERT-base models enhanced with PGKD (both CPU and GPU configurations), Claude-3 zero-shot, and LLaMA 3 8B zero-shot models.}
\label{tab:inference_cost_latency}
\end{table}
BERT-base + \gls{pgkd} processes a batch of 64 inference samples in 21.45 seconds on an AWS m5.4xlarge instance (16 vCPUs) for \$0.0046, and in 0.46 seconds on a g5.4xlarge GPU instance for \$0.01. Claude Sonnet inference takes 60.64 seconds per batch on average, costing \$0.38, for inputs averaging 1k tokens. Cheaper open-source models like LLaMA 3 8B take 58.06 seconds on average and cost \$0.06 per batch. \glspl{llm} inference for the proposed classification task is approximately 3X slower and 25X more expensive than BERT-base + PGKD on a CPU instance and approximately 130X slower and 6X more expensive on a GPU instance.

\section{Conclusion and Future Work}
Deploying \glspl{llm} in real-world text classification applications poses significant challenges due to high inference costs and latency. This research introduces \gls{pgkd}, a novel methodology designed to effectively distill \gls{llm} knowledge into faster, more efficient models for multi-class text classification. This study proves the substantial performance improvements of \gls{pgkd} distillation compared to regular fine-tuning of a BERT-base \gls{plm}, while maintaining limited inference latency and costs. Comparative analyses with existing knowledge distillation and augmentation strategies further underscore \gls{pgkd}'s practical performance enhancements. The proposed ablation study reveals the significant contributions of \gls{pgkd} components, such as Gradual Performance Checks on Validation Reports and Hard Negative Mining.
Cost and latency benchmarks provide compelling evidence of \gls{pgkd}'s efficiency. Compared to zero-shot \gls{llm} costs, BERT-base with \gls{pgkd} has proven to be significantly more cost-effective and up to 130X faster for a broad spectrum of multi-class classification tasks. 
Future research will investigate the impact of different teacher \glspl{llm} on the distillation process and explore the influence of the student model size on \gls{pgkd} effectiveness. Future research will also focus on exploring more advanced prompting techniques to optimize \gls{pgkd} performance.
While \gls{pgkd} is showcased for text classification tasks, its versatile framework can be extended to any \gls{llm} distillation task, including language generation, making it a powerful tool for optimizing performance across a wide range of \gls{ai} applications.

\newpage
\section{Limitations}
While \gls{pgkd} has demonstrated significant improvements in multi-class text classification tasks, there are some limitations to consider the following aspects.\\
\textbf{(1)~Dependence on \gls{llm} performance}: The effectiveness of \gls{pgkd} is inherently tied to the performance of the \gls{llm} used for knowledge distillation. If the \gls{llm} struggles with domain-specific data or fails to generate high-quality samples, it may limit the potential gains from the distillation process. Future work could explore the impact of using different \glspl{llm} or ensembles of \glspl{llm} to mitigate this limitation.\\
\textbf{(2)~Computational cost during distillation}: Although \gls{pgkd} results in a more efficient student model for inference, the distillation process itself can be computationally expensive due to the iterative generation of samples from the \gls{llm}. This may limit the scalability of the approach for vast datasets or frequent model updates.\\
\textbf{(3) Evaluation on a limited set of tasks}: While \gls{pgkd} has been evaluated on diverse datasets, the experiments are still limited to a specific set of multi-class text classification tasks. Further validation in a broader range of datasets, languages, and domains would strengthen the generalizability of the findings. Additionally, exploring the applicability of \gls{pgkd} to other \gls{nlp} tasks beyond classification, such as named entity recognition or question answering, could broaden its impact. This is something that will be addressed in future work.\\
\textbf{(4)~Sensitivity to prompt engineering}: The performance of \gls{pgkd} may be sensitive to the quality of the prompts used to guide the \gls{llm} in generating samples and poorly designed prompts could lead to suboptimal distillation results. Developing robust prompt engineering strategies or automating the prompt generation process could help mitigate this limitation and improve the consistency of \gls{pgkd}'s performance across different datasets and tasks.

Addressing these limitations could further enhance the applicability and robustness of the \gls{pgkd} methodology, making it an even more valuable tool for leveraging \glspl{llm} distillation in production-ready systems.

\section{Ethical Considerations}

\subsection{Benefits}

The proposed \gls{pgkd} methodology has significant societal implications, particularly in democratizing access to high-performance models in cost-constrained applications. By leveraging the knowledge distilled from \gls{llm}, \gls{pgkd} enables the development of more accurate and efficient models that can be deployed in a wide range of applications, including intent detection, topic classification, and other multi-class text classification tasks.

In many industrial and commercial settings, the high inference cost and latency of \gls{llm} can be a significant barrier to adoption. \gls{pgkd} offers a cost-effective and efficient solution, enabling organizations to leverage the power of \glspl{llm} without the associated costs and latency at inference time. This can profoundly impact the democratization of access to AI technology, particularly in resource-constrained environments.

Furthermore, \gls{pgkd} has the potential to benefit a wide range of industries and applications, including customer support, messaging platforms, and other areas where multi-class text classification is a critical component. By enabling the development of more accurate and efficient models, \gls{pgkd} can help to improve the overall quality of service and user experience in these applications.

\subsection{Potential Risks}

The development and application of \gls{pgkd} for multi-class text classification present potential fairness and bias considerations as \gls{pgkd}'s performance and the quality of student model outputs depends heavily on \gls{llm} output quality. If the teacher model for \gls{pgkd} contains bias or even worse hallucinates, it will generate biased and even hypothetical data points on which the student model will be trained. The distilled \gls{llm} knowledge may perpetuate and amplify existing student model biases and even impact training data quality. Ensuring the use of \glspl{llm} that are robust against bias and are well grounded in data generation is essential in \gls{pgkd} distillation.


\newpage
\appendix

\section{Prompt Details}
This section reports the prompts used in this work: `PGKD Prompt' is the prompt used for the Knowledge Distillation process.`Zero-shot Classification Prompt' is a prompt used for zero-shot classification with \gls{llm}. 
\label{sec:appendix_prompt}

\subsection{PGKD Prompt}

\begin{mdframed}[linewidth=1pt]
Human:
\newline
You are a Teacher model for a Student LM to perform topic detection on the following taxonomy: 
\newline
\{\textbf{Dataset Class Taxonomy}\}
\newline
Here are a few labeled examples that show the correct label for this task:
\newline
\{\textbf{Few-Shot Labeled Samples from Training Dataset}\}
\newline
Given the current model performance, please generate \{\textbf{PGKD Batch Size}\} training samples for the model to improve its performance. The response should be a list of dictionaries in JSON format, the response needs to be parsable so do not output anything else rather than the response itself. The objective is to maximize the model accuracy, generate new samples knowing that the classification report over validation set is:
\newline
\{\textbf{Classification Report on the Validation Set}\}
\newline
Please consider a few samples that the model was able to classify correctly:
\newline
\{\textbf{Correctly Classified Samples with correct label and Student-predicted label}\}
\newline
And samples the model was not able to classify correctly: 
\newline
\{\textbf{Misclassified Samples with correct label and Student-predicted label}\}
\newline
The model has a high confidence in classifying the following misclassified examples:
\newline
\{\textbf{Hard Negative Samples with correct label and Student-predicted label}\}
\newline 
Assistant:
\end{mdframed}

\subsection{Zero-shot Classification Prompt}

\begin{mdframed}[linewidth=1pt]
Human:
\newline
You are an AI assistant, and you are tasked to perform topic classification starting from text. You are asked
to classify text in topics categories. You are only
allowed to choose one of the
following categories:
\newline
\{\textbf{Dataset Class Taxonomy}\}
\newline
Please provide only
one category for each text in
JSON format.
For example: 
\newline
``class\_label'': ,
``class\_names'': ``''
\newline
Please
do not repeat or return the content
back again, just provide the
category in the defined format.
\newline
Text-to-classifiy: 
\newline
\{\textbf{Paste text for zero-shot classification}\}
\newline
Assistant:
\end{mdframed}

\section{Impact of Training Sample Size}
\label{appendix:scaling}
Table~\ref{tab:scaling_laws} illustrates the performance metrics of BERT-base and BERT-base augmented with \gls{pgkd} across different datasets as training samples increase. 
\begin{table*}[!ht]
\centering
\resizebox{\linewidth}{!}{%
\begin{tabular}{@{}lcccccc@{}}
\toprule
\multicolumn{1}{c}{\textbf{}} & \multicolumn{6}{c}{\textbf{Method}} \\
\cmidrule(l){2-7} 
\multicolumn{1}{c}{} & \multicolumn{3}{c}{BERT base} & \multicolumn{3}{c}{BERT base + PGKD} \\
\cmidrule(lr){2-4} \cmidrule(l){5-7}
Samples & Acc. & Macro Avg. F1 & Weighted   Avg. F1 & Acc. & Macro Avg. F1 & Weighted Avg. F1  \\
\midrule
\multicolumn{7}{c}{\textbf{AG-news (\#class = 4)}} \\
\hline
1000    & 0.884 & 0.884 & 0.884 & 0.895 & 0.894 & 0.894 \\
2500    & 0.899 & 0.899 & 0.899 & 0.910 & 0.909 & 0.909 \\
5000    & 0.903 & 0.903 & 0.903 & 0.911 & 0.911 & 0.911 \\
10000   & 0.914 & 0.914 & 0.914 & 0.917 & 0.917 & 0.917 \\
\midrule
\multicolumn{7}{c}{\textbf{Yahoo Answers (\#class = 10)}} \\
\hline
1000    & 0.649 & 0.657 & 0.657 & 0.685 & 0.688 & 0.688 \\
2500    & 0.700 & 0.693 & 0.693 & 0.708 & 0.705 & 0.704 \\
5000    & 0.702 & 0.697 & 0.697 & 0.719 & 0.707 & 0.709 \\
10000   & 0.721 & 0.716 & 0.716 & 0.725 & 0.718 & 0.721 \\
\midrule
\multicolumn{7}{c}{\textbf{Huffington Post (\#class = 41)}} \\
\hline
1000    & 0.474 & 0.214 & 0.411 & 0.519 & 0.330 & 0.495 \\
2500    & 0.537 & 0.343 & 0.519 & 0.553 & 0.394 & 0.541 \\
5000    & 0.551 & 0.378 & 0.547 & 0.564 & 0.425 & 0.564 \\
10000   & 0.601 & 0.387 & 0.561 & 0.605 & 0.476 & 0.604 \\
\midrule
\multicolumn{7}{c}{\textbf{AMZN Reviews (\#class = 335)}} \\
\hline
1000    & 0.320 & 0.074 & 0.244 & 0.443 & 0.159 & 0.382 \\
2500    & 0.541 & 0.243 & 0.491 & 0.602 & 0.340 & 0.567 \\
5000    & 0.648 & 0.404 & 0.622 & 0.666 & 0.436 & 0.649 \\
10000   & 0.687 & 0.451 & 0.662 & 0.694 & 0.476 & 0.679 \\
\bottomrule
\end{tabular}}
\caption{Comparative Performance of BERT-base and BERT-base + PGKD Across Different Datasets at Increasing Training Sample Sizes.}
\label{tab:scaling_laws}
\end{table*}
Noticeably, the improvement gap between \gls{pgkd} and BERT-base narrows with more training data. For instance, in the AG-news dataset, the performance difference in accuracy is 1.1\% with 1000 samples but decreases to just 0.3\% with 10,000 samples. This trend is consistently observed across all datasets, highlighting that while \gls{pgkd} enhances performance, its relative benefit diminishes as more training data is used. Importantly, \gls{pgkd} shows no signs of performance degradation, maintaining or improving upon the baseline metrics across all sample sizes and datasets.

\end{document}